\documentclass[10pt,twocolumn,letterpaper]{article}

\usepackage[pagenumbers]{cvpr} 
\usepackage{arydshln}
\definecolor{cvprblue}{rgb}{0.21,0.49,0.74}
\usepackage[pagebackref,breaklinks,colorlinks,allcolors=cvprblue]{hyperref}

\def\paperID{7487} 
\def\confName{CVPR}
\def\confYear{2025}

\def\ie{\emph{i.e.}\xspace}
\def\eg{\emph{e.g.}\xspace}
\newcommand{\mypartitle}[2][2.5]{\vspace*{-#1 ex}~\\{\noindent {\bf #2}}}

\newcommand{\Gazepseudo}{\ensuremath{\mathbf{g}^{ps}}\xspace}

\newcommand{\Gazepredicted}{\ensuremath{\hat{\mathbf{g}}}\xspace}
\newcommand{\Gazepredictedcoordinate}{\ensuremath{\hat{g}}}

\newcommand{\Inputpatch}{\ensuremath{\textbf{x}}\xspace}

\newcommand{\model}{GaT\xspace}
\newcommand{\framework}{ST-WSGE\xspace}
\newcommand{\papertitle}{Enhancing 3D Gaze Estimation in the Wild using\\ Weak Supervision with Gaze Following Labels}

\author{Pierre Vuillecard\\
IDIAP, EPFL\\
{\tt\small pierre.vuillecard@idiap.ch}
\and
Jean-Marc Odobez\\
IDIAP, EPFL\\
{\tt\small odobez@idiap.ch}
}

\begin{document}

\title{ \papertitle }

\twocolumn[{%
\renewcommand\twocolumn[1][]{#1}%
\maketitle

\vspace{-1.2cm}
\begin{center}
    \centering
    \captionsetup{type=figure}
    \includegraphics[width=1.\textwidth]{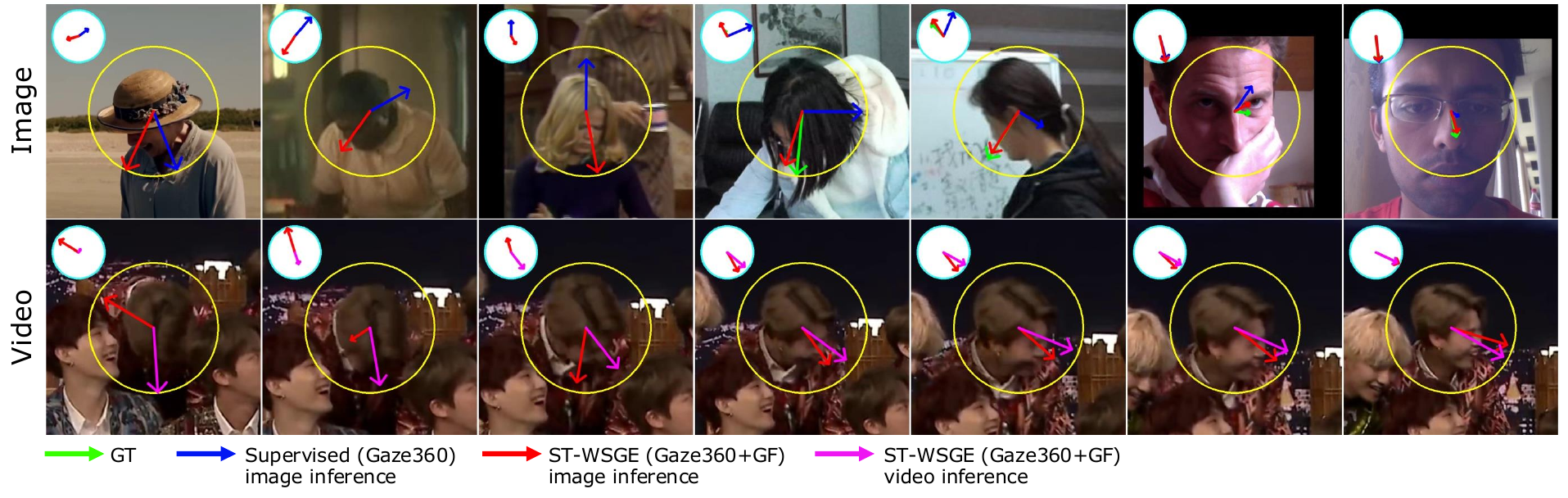}
    \vspace{-6mm}
    \captionof{figure}{
    \textbf{Significance of ST-WSGE.} Our self-training based weakly-supervised framework for robust 3D gaze estimation in real-world conditions (e.g., varying appearance, extreme poses, resolution, and occlusion).  
     All predictions used our image and video agnostic Gaze Transformer (\model) model.
     Top row: importance of the training diversity using ST-WSGE and GazeFollow (GF) for generalization compared to standard supervised methods. 
     Bottom row: influence of temporal context between image and video inference.  Circles in  images represent unit disks where 3D gaze vectors are projected onto the image plane (x, y in yellow) and a top-down view (x, z in blue). Images from VideoAttentionTarget, GFIE, and MPIIFaceGaze datasets.
}
    \label{fig:intro}
\end{center}%
}]


\begin{abstract}
Accurate 3D gaze estimation in unconstrained real-world environments remains a significant challenge due to variations in appearance, head pose, occlusion, and the limited availability of in-the-wild 3D gaze datasets. To address these challenges, we introduce a novel Self-Training Weakly-Supervised Gaze Estimation framework (\framework). This two-stage learning framework leverages diverse 2D gaze datasets, such as gaze-following data, which offer rich variations in appearances, natural scenes, and gaze distributions, and proposes an approach to generate 3D pseudo-labels and enhance model generalization.
Furthermore, traditional modality-specific models, designed separately for images or videos, limit the effective use of available training data. To overcome this, we propose the Gaze Transformer (\model), a modality-agnostic architecture capable of simultaneously learning static and dynamic gaze information from both image and video datasets. By combining 3D video datasets with 2D gaze target labels from gaze following tasks, our approach achieves the following key contributions:
(i) Significant state-of-the-art improvements in within-domain and cross-domain generalization on unconstrained benchmarks like Gaze360 and GFIE, with notable cross-modal gains in video gaze estimation;
(ii) Superior cross-domain performance on datasets such as MPIIFaceGaze and Gaze360 compared to frontal face methods.
Code and pre-trained models will be released to the community.
\end{abstract}
\vspace{-0.5cm}



\section{Introduction}
\label{sec:intro}

\noindent Non-verbal behaviors play a crucial role in human communication, often conveying more information than words alone. Among the various forms of non-verbal cues, eye gaze stands out as an important signal for understanding human behavior, including attention, communication, intents, and mental state. 
Consequently, gaze signals have been used in many applications. Some applications require accurate gaze for frontal head pose such as AR/VR \cite{burova2020utilizing}, 3D avatar animation \cite{richard2021audio}, human-computer interaction \cite{andrist2014conversational}, and driver behavior monitoring \cite{kasahara2022look}. While other applications focus on robust 3D gaze estimation from a wide range of head poses such as medical and psychological analysis \cite{kleinke1986gaze} or human-robot interaction \cite{admoni2017social}. 

In this paper, our goal is to develop a robust 3D gaze estimation for in-the-wild applications with unconstrained head pose and real-world environments. In the literature, this refers to the less explored and highly challenging problem of ``physically unconstrained gaze estimation"~\cite{Gaze360}.

\mypartitle{Motivations.}
Estimating gaze in unconstrained, real-world settings poses unique challenges not fully addressed by current lab-based datasets, which are primarily collected in controlled screen-target setups \cite{Rt-gene, GazeCapture, MPIIGaze, Eyediap, EVE, ETH}. While these datasets have enabled recent approaches to achieve high accuracy in frontal 3D gaze estimation from monocular images \cite{zhang2017s, chen2018appearance, Rt-gene, cheng2020coarse, cheng2022gaze, abdelrahman2023l2cs, yan2023gaze, catruna2024crossgaze}, their effectiveness declines in real-world scenarios. This limitation stems from restricted gaze distributions, lab-specific conditions, limited subject diversity, and reliance on potentially noisy head-pose estimates for normalizing eye and face images~\cite{zhang2018revisiting}.

To address the lack of data for unconstrained gaze estimation, Gaze360 \cite{Gaze360} and GFIE \cite{GFIE} were developed. Although these datasets have advanced the field, models trained on them continue to struggle with challenging, real-world conditions (see \cref{fig:intro}), particularly when facing extreme head poses, partial eye occlusions, varying resolutions from diverse camera-to-subject distances, and a wide range of appearances (e.g., skin tones, hairstyles, facial expressions). This limitation is largely due to insufficient diversity in the training data, as collecting high-quality, naturally occurring, and diverse 3D gaze data is complex, resource-intensive, and not easily scalable.

To overcome these limitations, researchers have explored using “secondary” labels that are easier to obtain, \eg by relying on internet data. 
For example, Kothari \etal~\cite{kothari2021weakly} leveraged 2D gaze direction labels from the “Looking at Each Other” (LAEO) dataset~\cite{marin2019laeo}. By applying geometric constraints and head-size heuristics, they generated pseudo-3D gaze data.

While this approach showed some generalization improvement, the authors noted that LAEO's gaze distribution is primarily horizontal and requires images containing at least two people with mutual gaze, which limits sample diversity and availability.

Here we aim to utilize more general 2D gaze annotations from the gaze following task \cite{recasens2015they, Chong_2018_ECCV}. Although the ground truth for gaze following is defined as the 2D pixel location a person in the scene is looking at, we can repurpose it as 2D gaze direction ground truth. Compared to LAEO, gaze following datasets offer greater diversity, with broader gaze distributions and a wider variety of natural scenes.

In addition, in contrast to \cite{kothari2021weakly}, we propose a Self-Training Weakly-Supervised Gaze Estimation (\framework) framework using a two-stage training approach without relying on heuristics or relative depth estimation to generate pseudo-3D gaze labels. First, we train a gaze network on existing 3D gaze datasets. We then use this network's predictions on gaze-following data, combined with 2D gaze ground truth, to create 3D gaze pseudo-labels. In the second stage, we retrain a similar gaze network using both 3D gaze datasets and gaze-following datasets with these pseudo-labels.
Our approach minimizes the need for labor-intensive, unconstrained 3D gaze labeling and demonstrates significant improvements over state-of-the-art methods in both within-domain and cross-domain generalization on Gaze360, GFIE, and MPIIFaceGaze \cite{MPIIGaze}.

Given the scarcity of in-the-wild 3D gaze datasets, 
another challenge lies in how to leverage both image and video data effectively. 
Modality-specific models restrict the training set to modality-specific datasets, limiting their ability to benefit from all available resources. While static models can draw on large image datasets such as GazeFollow \cite{Chong_2018_ECCV}, temporal dynamics are also essential for robust 3D gaze estimation in unconstrained environments \cite{Gaze360, nonaka2022dynamic}, 
and is particularly valuable when the eye region is obscured,
whether due to occlusions, low resolution, or blinking (see \cref{fig:intro}).

To address this, one approach is to pre-train a model on images and transfer the weights to a temporal model using techniques like filter inflation, where 2D filters are extended to 3D models, as done in prior adaptations for video tasks \cite{carreira2017quo}. However, this method is more suited to fine-tuning and risks catastrophic forgetting, where the model loses pre-trained knowledge~\cite{mccloskey1989catastrophic}.
Alternatively, images can be duplicated to simulate fixed-length video clips, allowing for training on both image and video datasets in a temporal model. However, this can impair the learned gaze dynamics, as synthetic clips lack genuine motion information.

Transformers offer a promising solution for handling multiple modalities. Inspired by recent work \cite{girdhar2022omnivore, girdhar2023omnimae}, we propose a Gaze Transformer (\model) designed to encode both image and video inputs into a shared representation. 
This allows us to leverage labeled datasets more effectively,
by training jointly on image and video data. 
We  demonstrate better cross-modal generalization, 
and that using image datasets enhances video gaze prediction, thus enabling a more versatile and robust 3D gaze estimation model.

\mypartitle{Contributions.}
They can be summarized as:
\begin{itemize}
\item {\bf \framework, a novel learning framework enhancing generalization.} 
To address the lack of diverse, naturalistic 3D gaze datasets, we leverage 2D gaze-following datasets using 3D pseudo labels. Combining these with 3D gaze datasets in a two-stage manner, we demonstrate improved 3D generalization on several benchmarks.
\item {\bf A visual modality agnostic Gaze Transformer (\model) architecture 
making efficient use of existing gaze datasets.}
By allowing simultaneous learning from 3D gaze image and video datasets, it outperforms modality-specific models, resulting in better static and dynamic gaze representations, better capturing spatiotemporal patterns in head sequences compared to the state-of-the-art.
\item {\bf State-of-the-art results.} 
Our approach surpasses existing methods in both unconstrained (Gaze360, GFIE) and constrained (MPIIFaceGaze) environments, achieving superior results in within- and cross-dataset evaluations.
\end{itemize}
These contributions position our approach as ideal for real-world unconstrained 3D gaze estimation applications.

\vspace{-1mm}
\section{Related Work}
\vspace{-1mm}
Our research pertains to 3 main aspects:
unconstrained gaze estimation, temporal gaze modeling, 
and generalization using additional data and labels to bridge the domain gap between controlled setups and real-world data.\\
\mypartitle{Unconstrained Gaze Estimation.} 
%
Most 3D gaze estimation models address the frontal face gaze prediction task 
\cite{zhang2017s,chen2018appearance,Rt-gene,cheng2020coarse, cheng2022gaze,abdelrahman2023l2cs,yan2023gaze}, 
relying on normalized frontal face crop as input. 
These methods tend to fail under partial occlusion of the eyes due to extreme head pose. Nevertheless, at 90-135 head pose yaw, a significant part of one eyeball is still often visible and informative for gaze estimation \cite{Gaze360}.
For this reason, few works tackle the most challenging setting of
``physically unconstrained gaze estimation" without constraint on the head pose. 
Kellnhofer \etal~\cite{Gaze360} are the first to collect a physically unconstrained 3D gaze dataset Gaze360 and develop a method that used head crop as input. Then, combining different head crop scales proved to be beneficial \cite{chen2020360} since more resolution helps on the frontal face while more context is beneficial for profiles and back heads. Following this idea, MCGaze \cite{guan2023end} used a spatiotemporal interaction module between head, face, and eye features in an end-to-end manner to extract local eyes and global head features. These approaches focus on within-data performance, while in this work we aim to improve both within and the generalization as discussed in the following section. \\
\mypartitle{Generalization in the Wild.} Bridging the dataset's domain gap challenge is crucial for 3D gaze estimation in real-world applications. 
Two trends have been explored to adapt to specific target domains effectively: One leverages few labeled samples, while the other uses only unlabeled samples~\cite{liu2021generalizing,cheng2022puregaze,wang2022contrastive,ververas20223dgazenet,Gaze360,jindal2024spatio}.
In contrast, gaze generalization models focus on enhancing cross-domain performance without any prior knowledge of the target domain. For instance, the methods proposed in \cite{cheng2022puregaze,bao2022generalizing,wang2022contrastive} demonstrate improved generalization by learning robust general features (\eg via image rotation consistency) for gaze estimation across varying conditions. Even if those methods focus on constrained settings with face crop as input, we compare our approach with them to show the effectiveness in frontal face generalization.\\
Furthermore, to improve in-the-wild generalization, researchers seek to exploit diverse weak gaze labels that can be easily or automatically generated on in-the-wild data. 
In this direction, Zang \etal~\cite{MPS} automatically generates a new 3D gaze dataset, MPSGaze, by blending on images of people from the  Widerface datasets~\cite{yang2016wider}, eyes from images of the  ETH-Xgaze dataset with known 3D gaze and similar head pose. 
While this greatly improves diversity with more than 10000 new identities, this method generates only near frontal faces and might impact the appearance of the face. 
In another study, Ververas \etal \cite{ververas20223dgazenet} used eyeball fitting techniques to create pseudo-3D gaze on new face datasets. They improved generalization, but their work is also restricted to frontal faces. 
Finally, Kothari \etal \cite{kothari2021weakly} used a weakly-supervised learning framework for improved generalization using pseudo 3D gaze labels from 2D gaze LAEO labeled datasets. 
However, as acknowledged by the authors, the 2D gaze distribution of LAEO is limited horizontally. 
In our work, we follow this idea but leverage a more diverse gaze distribution and natural scene 2D gaze label obtained from the annotation of where people look in the scene. Using a self-training learning approach with generated 3D pseudo labels via geometric projection, we show improved within and cross-dataset generalization on unconstrained Gaze360 and GFIE \cite{GFIE} and frontal MPIIFaceGaze~\cite{MPIIGaze} datasets.\\
\begin{figure*}[htbp]
    \centering
    \includegraphics[width=\linewidth]{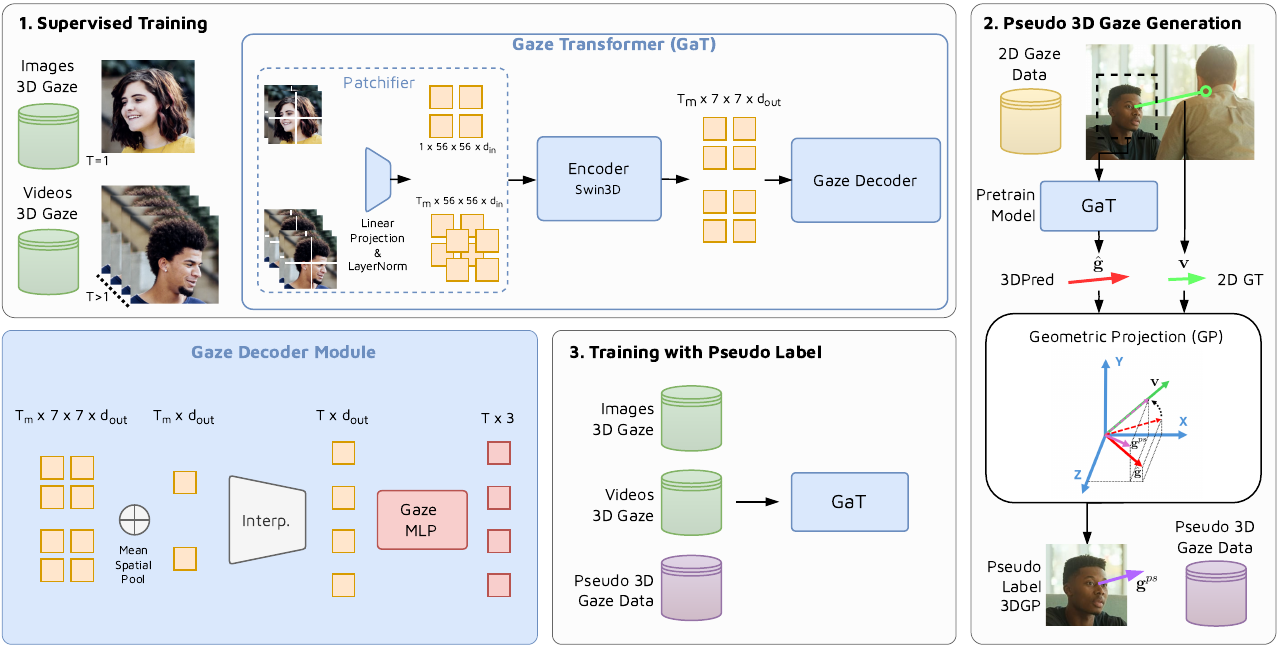}
    \vspace{-2mm}
    \caption{
    \textbf{Our \framework training framework.}
    1. In the first stage, we train a Gaze Transformer (\model) on both image and video 3D gaze datasets.
    2. Using the trained network, 3D gaze is inferred on 2D gaze dataset. Then, a geometric rotation is applied to generate a pseudo 3D gaze label from the inferred 3D gaze that is aligned to the 2D ground truth gaze label in the image plane.
    3. In the second stage, we train a similar gaze network as in 1. using available 3D gaze datasets and gaze following datasets with 3D pseudo labels.
    }
    \label{fig:approach}
    \vspace{-4mm}
\end{figure*}
\mypartitle{Dynamic 3D Gaze Estimation} 
has not been extensively explored due to the lack of available datasets. 
EYEDIAP~\cite{Eyediap} and EVE \cite{EVE} are video datasets collected in constrained settings, resulting in mostly frontal poses. 
In this particular context, Park \etal~\cite{EVE} and Palmero \etal~\cite{palmero2018recurrent} estimated the gaze from face crop image sequences but only showed marginal improvement compared to static methods. Indeed, it is questionable if eyeball dynamics have temporal dependencies besides the ones due to specific tasks or scenarios (\eg reading). 
Nevertheless, in unconstrained settings with low resolution and head pose dynamic scenarios, temporal methods show benefits in encoding the head and eye dynamics \cite{Gaze360,nonaka2022dynamic,chen2020360,guan2023end}. 
For instance, seen from a far distance, head and body orientation dynamic 
revealed to be an important prior for gaze estimation when eyes are barely visible \cite{nonaka2022dynamic}.
Unconstrained video gaze data is challenging to collect. Beyond Gaze360, GFIE \cite{GFIE} is the only other 3D gaze dataset for gaze following, using complex laser setups, yet it is limited to indoor settings and lacks natural scene and gaze dynamics. The scarcity of video 3D gaze datasets hampers the development of video-based methods that generalize to real-world data. To address this, we propose a unified model trained on both image and video datasets, demonstrating improved video prediction through diverse, large-scale data. \\
Furthermore, current video-based gaze estimation methods typically employ a backbone to extract features from image sequences, followed by a Recurrent Neural Network (RNN) to capture temporal dynamics \cite{Gaze360,EVE,palmero2018recurrent,jindal2024spatio,chen2020360,kothari2021weakly}. However, these approaches do not explicitly model the spatiotemporal interactions in the input sequence. To address this limitation, we investigate a spatio-temporal model to encode subtle eye motion or head pose changes in the input sequence directly. \\

\vspace{-2mm}
\section{\framework Method}
\vspace{-1mm}

\subsection{Self-Training Pipeline}
\label{sec:self_taining}

\vspace{-1mm}

We propose a two-stage training pipeline for gaze estimation to leverage any 2D gaze datasets, as presented in \cref{fig:approach}. 
In the first stage, a gaze network is trained on image and video 3D gaze datasets in a supervised manner. Next, the network is used to infer 3D gaze on 2D gaze datasets. 
Since only the gaze's depth is missing in the 2D gaze label, we employ a geometric transformation to generate a robust pseudo-3D gaze label from the inferred 3D gaze that is aligned with the 2D gaze label. We assume that a model pre-trained on unconstrained 3D gaze datasets provides a good prior z-estimation. In the second stage, a similar gaze network is trained in a supervised regime using both gaze following data with 3D pseudo labels and 3D gaze datasets. 

\vspace{-1mm}
\subsection{ Gaze Transformer (\model)}
\label{sec:model}
\vspace{-1mm}

\mypartitle{Model Architecture Motivation.}
Accurate and robust 3D gaze estimation in the wild requires three key capabilities: capturing fine local details from the eye region when visible; extracting global information from head orientation which is particularly useful when the eyes are occluded or partially obfuscated; capturing small motion of head pose and eyes in the temporal domain to capture subtle gaze shifts. A model capable of training on both image and video data is especially valuable, as it broadens the range of available training datasets. 
Convolutional Neural Networks (CNNs) excel in frontal gaze estimation due to their ability to extract local eye features \cite{Gaze360,abdelrahman2023l2cs,yan2023gaze} but may encounter more difficulty in global reasoning (i.e. merging pose and eye information), and extending CNNs to handle temporal data within a single modality-agnostic model is challenging.
Vision Transformers (ViTs) \cite{dosovitskiy2020image}, as noted by Cheng \etal~\cite{cheng2022gaze}, are less suited for gaze estimation since they may miss critical local details, especially when the eye region is split across multiple patches.
In contrast, hierarchical transformer architecture \cite{liu2021swin,ryali2023hiera} offers a flexible architecture to capture both local and global features. For instance, the Swin Transformer, which uses smaller patches (typically $4\times4$ vs $16\times16$ in standard ViTs), is better able to capture fine local details. Its "shifted window" mechanism, which applies self-attention within local windows that shift at regular intervals, effectively aggregates local and global context. Extending the Swin Transformer to the temporal dimension has proven successful for temporal tasks on several benchmarks~\cite {liu2022video}.
Additionally, transformers are versatile, recent work has demonstrated their effectiveness when trained on both image and video datasets within a single model~\cite{girdhar2022omnivore, girdhar2023omnimae}. Inspired by these approaches, we introduce our Gaze Transformer, \model, with several modifications for 3D gaze estimation, as illustrated in \cref{fig:approach} and detailed in the following sections.\\
\mypartitle{Patchifier.}
The model needs a common representation format to encode both image and video input. 
Following \cite{dosovitskiy2020image,liu2021swin,feichtenhofer2022masked,tong2022videomae,girdhar2022omnivore}, images and videos are represented as 4D tensors $\mathrm{\textbf{X}} \in \mathbb{R}^{T \times H\times W \times 3}$, 
with  $T=1$ for an image $\mathrm{\textbf{I}}$,  and $T>1$ for a video clip $\mathrm{\textbf{V}}$.
Then, the input $\mathrm{\textbf{X}}$ is divided into a collection $\{ \Inputpatch_i \}^N$ of 4D sub-tensor (patches) 
$\Inputpatch_i \in \mathbb{R}^{\mathrm{t} \times \mathrm{h} \times \mathrm{w} \times 3}$,  as presented in \cref{fig:approach}. 
Following \cite{girdhar2022omnivore,tong2022videomae, girdhar2023omnimae,marin2019laeo}, we use $t=2$.
When working with image only,  we duplicate the image instead of zero-padding because we find better cross modalities generalization from video to image. Then, a shared linear layer and LayerNorm are applied to project the patches to a token representation. \\
\mypartitle{Encoder.}
The tokens from the patchifier are then fed into a tiny Swin3D hierarchical spatiotemporal encoder. It relies on self-attention within nearby tokens in a spatiotemporal window that is shifted every time. In addition, it uses two sets of relative positional encoding: one spatial and one temporal. Because of the hierarchical representation, the number of tokens is reduced by patch merging layers as the network
gets deeper. The temporal output dimension is reduced by a factor of two. The output tokens are then fed to a gaze decoder module.  \\
\mypartitle{Gaze Decoder.}
We first apply a mean spatial pooling on the output tokens, followed by an interpolation function to double the temporal dimension to match the input length (for images, interpolation is skipped). Finally, a shared MLP with a single hidden layer is applied to each token to predict a normalized 3D gaze vector. \\
\mypartitle{Baseline Networks.} 
Different approaches exist to process image and video in a single model. To compare the performance of our \model model, we develop in addition two baselines.
Given that the static Swin(2D) transformer gives good performance on gaze estimation. We add a temporal encoder to model the gaze dynamic. Therefore, we develop Swin(2D)-LSTM which first processes a set of images using Swin and outputs a set of embedding for each image. Then, it is fed to a bidirectional LSTM followed by a shared gaze MLP on each output to produce a gaze vector. 
Similarly, the second baseline called Swin(2D)-Tr replaces the LSTM by a transformer. The output tokens from each image are projected to a lower dimension followed by a LayerNorm and absolute spatiotemporal encoding. 
Then, the transformer is applied to the spatiotemporal output token. Finally, a spatial mean pooling is applied followed by a similar gaze MLP. Both architectures are input agnostic and are compared in an ablation study.

\subsection{Pseudo 3D Gaze Generation }
\vspace{-1mm}
Creating 2D gaze datasets is easier than creating 3D gaze datasets, as annotation can be completed after the images have been collected, unlike 3D gaze which can not be annotated by humans and require a special setup. 
As a result, 2D datasets like GazeFollow offer a broad gaze, head pose distribution, and head/face appearance diversity. 
There are different possibilities to leverage such a dataset which will be discussed in the ablation sections.  
For instance, Kothari \etal~\cite{kothari2021weakly} used 2D gaze from LAEO labels and 3D fitted head models for z-direction estimation. 
In contrast, as presented in \cref{fig:approach}, our 3D pseudo gaze generation method assumes that a pre-trained model trained on unconstrained 3D gaze datasets can provide a good prior z-estimation, which is confirmed by our experiments. Combining the z component of predicted 3D gaze with 2D gaze ground truth 
provides a robust pseudo 3D gaze label. 
Then, using this pseudo label as an additional label during training, we report improvement in unconstrained generalization. \\
\mypartitle{Geometric Projection.}
The predicted 3D Gaze (3DPred) $\Gazepredicted$ and the 2D ground truth (2D GT) $\textbf{v} = (v_x,v_y)$ are combined such that the image plane projection of the pseudo 3D gaze (3DGP) $\Gazepseudo$ is aligned with the 2D ground truth $\textbf{v}$. 
Therefore, a rotation is applied to $\Gazepredicted$ around the z-axis such that $\Gazepseudo$ has the same x,y direction as $\textbf{v}$:
\vspace{-0.1cm}
\begin{equation}
    \Gazepseudo = ( v_x \|(\Gazepredictedcoordinate_x,\Gazepredictedcoordinate_y) \|_2, v_y \|(\Gazepredictedcoordinate_x,\Gazepredictedcoordinate_y) \|_2, \Gazepredictedcoordinate_z)
\end{equation}
\vspace{-6mm}
\subsection{Training Strategy}
\vspace{-1mm}

In both training stages, the objective is to train our \model model on a collection of both image and video datasets with gaze label 
$\{(\mathrm{\textbf{X}},\textbf{g})_j\}$ where $\textbf{g} \in \mathbb{R}^{T \times 3}$ with T=1 for images, which creates different training challenges to be addressed.
\mypartitle{Video Training Data.}
As our model is modality agnostic, a video dataset can be 
considered both as a set of video clips or as a collection of images.
These views of the data are not equivalent, as, typically, considering the data as video-clip at training will impact more inference on videos at test time rather than on images. Hence, a video dataset can be used twice as an image or video training dataset\footnote{By convention, when reporting experiments, for video datasets, we will add the suffix I when it is considered as an image dataset, V in the video case, and I\&V when it is used twice as image and as video dataset.}.
We will see in ablations that it can impact the modality generalization capability.

\mypartitle{Mini-batch Strategy.}
Different mini-batch strategies have been proposed in the literature to handle multiple datasets.
One approach mixes samples from each of the datasets, but in our case, this requires careful implementation because images and videos don't have the same dimensions.
Another strategy creates batches from one dataset at a time and alternates between them. This approach has proven effective in previous work \cite{girdhar2022omnivore,hu2021unit} and we followed this approach here. 
In addition, dataset size imbalance is another challenge, as dataset sizes range from 30k to 120k samples. To address this, we balance the datasets by oversampling smaller ones and undersampling larger ones so that each dataset contributes equally during an epoch.\\
\mypartitle{Loss.}
For training the model, we utilize a temporal weighted average of the angular loss, which represents the angular difference between the predicted gaze vector $\hat{\textbf{g}}$ and the ground truth ${\textbf{g}}$ in degree:\\[-3mm]
\begin{equation}
    \mathcal{L}_{gaze}(\hat{\textbf{g}}, \textbf{g}) = \frac{1}{T}\sum^T_{t=1}\frac{180}{\pi}\arccos(\frac{\hat{\textbf{g}}_t^T\textbf{g}_t}{\|\hat{\textbf{g}}_t\| \|\textbf{g}_t\|})
\end{equation}

\begin{table}[t]
\centering
\resizebox{\columnwidth}{!}{%
\begin{tabular}{llccccrrrrrrrrc}
\toprule
 &  & \multicolumn{4}{c}{GF Additional Label} & \multicolumn{2}{c}{G360 Full}  & \multicolumn{1}{l}{} & \multicolumn{2}{c}{GFIE} & \multicolumn{1}{l}{} & \multicolumn{1}{c}{MPII}& \multicolumn{1}{l}{} & \multicolumn{1}{c}{EDIAP} \\ 
 \cline{7-8} \cline{10-11} \cline{13-13}  \cline{15-15}
Method & Training Data & None & 2D & 3DPred & 3DGP  & Img & Vid &  & Img & Vid & & Img & &  Vid \\ 
\midrule
Supervised &  G360I\&V & \checkmark & & &  & 13.6 & 12.6 &   & 21.9 & 20.9 &   & 7.4 &   & 8.3 \\
Weakly-Sup (WS)&  G360I\&V+GF & & \checkmark & &  & \textbf{13.1} & \textbf{12.1} &   & \underline{16.1} & \underline{15.7} &   & \underline{6.5} &   &  9.2 \\
Self-Training (ST)&  G360I\&V+GF & & & \checkmark &  & 13.6 & 12.7 &   & 20.2 & 19.7 &   & 7.4 &   &  \textbf{7.7} \\
\framework &  G360I\&V+GF & & &  & \checkmark & \underline{13.2} & \underline{12.2} &   & \textbf{15.9} & \textbf{15.5} &   & \textbf{6.4} &   & \underline{8.2} \\
\midrule
Supervised &  GFIEI\&V & \checkmark & & &  & 30.6 & 29.9 &   & 15.7 & 15.4 &   & 23.8 &   & 37.8 \\
Weakly-Sup (WS)&  GFIEI\&V+GF & & \checkmark & &  & 22.9 & 22.1 &   & \textbf{12.5} & \textbf{12.2} &   & 24.4 &   & 33.0 \\
Self-Training (ST)&  GFIEI\&V+GF & & & \checkmark &  & 29.7 & 29.4 &   & 14.9 & 14.9 &   & 21.2 &   &  34.6 \\
\framework &  GFIEI\&V+GF & & &  & \checkmark & \textbf{21.5} &\textbf{ 21.1} &   & \underline{13.0} & \underline{12.7} &   & \textbf{17.3} &   &  \textbf{16.7} \\
\bottomrule
\end{tabular}%
}
\vspace{-2mm}
\caption{\textbf{Ablation study for the self-training weakly-supervised learning framework.} We experiment with our \model model, two 3D gaze datasets Gaze360 and GFIE, and three ways to include GazeFollowing labels (GF). The \textbf{best} and the \underline{second} best scores are in bold and underlined, respectively.}
\label{tab:ablation_framework}
\vspace{-2mm}
\end{table}


\begin{table}[t]
\centering
\resizebox{\columnwidth}{!}{%
\begin{tabular}{lccrrrrrrrrr}
\toprule
  & \multicolumn{2}{l}{Training modality} & \multicolumn{2}{c}{G360 Full} & \multicolumn{1}{l}{} & \multicolumn{2}{c}{G360 180} & \multicolumn{1}{l}{} & \multicolumn{2}{c}{G360 40}  \\ 
 \cline{4-5} \cline{7-8} \cline{10-11}
Model  & Img & Vid & Img & Vid &  & Img & Vid &  & Img & Vid \\ 
\midrule
Swin(2D)-LSTM &  \checkmark &   & 14.33 & 13.97 &   & 12.17 & 11.89 &   & 9.73 & 9.47     \\
Swin(2D)-LSTM &     & \checkmark  & 14.75 & 13.05 &   & 12.58 & 10.98 &   & 9.86 & 8.67     \\
Swin(2D)-LSTM &  \checkmark  &  \checkmark & 13.93 & 13.02 &   & 11.76 & 10.94 &   & 8.88 & 8.27    \\
\midrule
Swin(2D)-Tr &   \checkmark  &  & 14.05 & 14.53 &   & 12.02 & 12.78 &   & 9.63 & 9.32  \\
Swin(2D)-Tr &     & \checkmark  & 14.05 & 12.63 &   & 12.05 & 10.54 &   & 9.46 & 8.14  \\
Swin(2D)-Tr &  \checkmark  &  \checkmark & 13.81 & 12.91 &   & 11.96 & 11.09 &   & 9.43 & 8.67  \\
\midrule
\model  &   \checkmark &   & 13.95 & 13.82 &   & 11.95 & 11.78 &   & 9.58 & 8.89  \\
\model  &   &  \checkmark & 13.87 & 12.31 &   & 11.89 & 10.39 &   & 9.29 & 7.95  \\
\model  &  \checkmark  & \checkmark  & 13.64 & 12.60 &   & 11.66 & 10.67 &   & 9.10 & 8.23  \\
\bottomrule
\end{tabular}%
}
\vspace{-2mm}
\caption{ \textbf{Ablation study for the gaze model network.} Since different models are image and video training agnostic, we experiment with three models on three training modalities dataset combinations using Gaze360 as the training set.}
\label{tab:ablation_model}
\vspace{-2mm}
\end{table}

\begin{table}[t]
\centering
\resizebox{\columnwidth}{!}{%
\begin{tabular}{lrrrrrrrrrrr}
\toprule
 &  \multicolumn{2}{c}{G360 Full} & \multicolumn{1}{l}{} & \multicolumn{2}{c}{G360 180} & \multicolumn{1}{l}{} & \multicolumn{2}{c}{G360 40} & \multicolumn{1}{l}{} & \multicolumn{2}{c}{GFIE}  \\
 \cline{2-3} \cline{5-6} \cline{8-9} \cline{11-12}  
Training Data & Img & Vid &  & Img & Vid &  & Img & Vid &  & Img & Vid  \\ 
\midrule
G360V+GF    & 13.5 & 12.1 &   & 11.6 & 10.2 &   & 8.3 & 7.7 &   & 15.7 & 17.9 \\
G360I\&V+GF    & 13.2 & 12.2 &   & 11.3 & 10.3 &   & 8.6 & 7.7 &   & 15.9 & 15.5  \\
\midrule
GFIEV+GF  & 22.8 & 24.2 &   & 22.4 & 23.9 &   & 29.9 & 31.8 &   & 13.4 & 13.0  \\
GFIEI\&V+GF  & 21.5 & 21.1 &   & 20.6 & 20.3 &   & 26.6 & 26.7 &   & 13.0 & 12.7 \\
\bottomrule
\end{tabular}%
}
\vspace{-2mm}
\caption{ \textbf{Impact of the training datasets modalities on cross-modal generalization.} We experiment with \model model, \framework framework, and different training dataset modalities. }
\label{tab:ablation_cross_modal}
\vspace{-2mm}
\end{table}

\vspace{-4mm}
\section{Experiments}
\vspace{-1mm}

\subsection{Datasets}

\begin{figure}[t]
    \centering
    \includegraphics[width=0.9\linewidth]{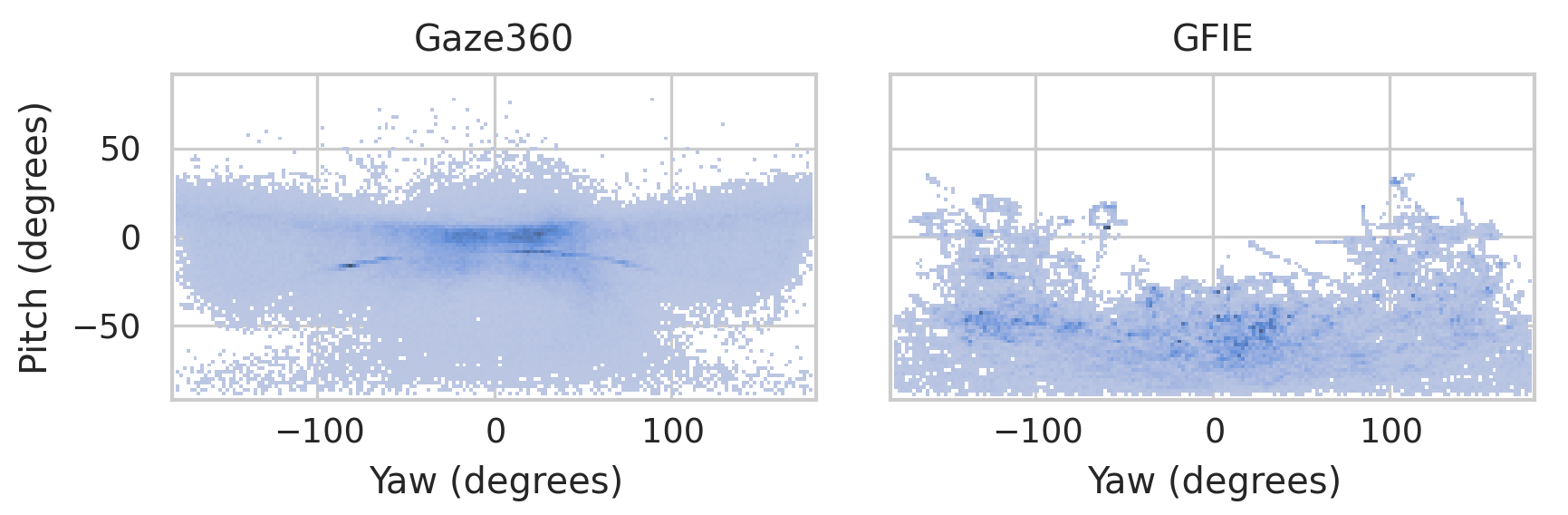}
    \vspace{-2mm}
    \caption{\textbf{Dataset gaze distribution.} Gaze in polar coordinates.}
    \label{fig:data_distribution}
    \vspace{-4mm}
\end{figure}

In this work, we employ five 3D gaze datasets:
two video unconstrained datasets for training and evaluation: Gaze360 (G360)~\cite{Gaze360}, GFIE~\cite{GFIE}, and three constrained only for generalization MPSGaze (MPS)~\cite{MPS}, MPIIFaceGaze (MPII)~\cite{MPIIGaze} and EYEDIAP (EDIAP)~\cite{Eyediap} (EDIAP), with only EDIAP being a video dataset. 
As shown in \cref{fig:data_distribution}, G360 and GFIE differ considerably in their gaze distribution, which makes cross-dataset evaluations challenging. 
In addition, we consider the 2D gaze following dataset GazeFollow \cite{recasens2015they} (GF), which contains more than 100k images with gaze target annotations.\\
The details of the six datasets are presented in the supplementary materials. 
Nevertheless, as authors have been using many subsets of G360 for evaluation, we clarify the test splits to avoid any confusion. 
We followed the split of \cite{Gaze360}: G360 Full corresponds to "All 360°" (all the test set); G360 180 corresponds to "Front 180°" (gaze within 90°); and G360 40 to "Front Facing" (gaze within 20°). Additionally, we consider G360 Back (gaze above 90°)~\cite{chen2020360} and G360 Face (all detected faces), used in many studies~\cite{zhang2017s,chen2018appearance,Rt-gene, cheng2020coarse, cheng2022gaze, abdelrahman2023l2cs, yan2023gaze, catruna2024crossgaze}. G360 Face 180 or 40 corresponds to the detected face with a gaze within 90° or 20°.

\vspace{-2mm}
\subsection{Implementation Details}
\vspace{-1mm}

Each dataset has different head bounding boxes ground truth. 
To avoid discrepancies in cropping, 
we standardize the input by using a robust pre-trained head detector\footnote{https://github.com/zhangda1018/yolov5-crowdhuman} train on the CrowdHuman dataset~\cite{shao2018crowdhuman}. 
We match the detected and ground truth bounding boxes to get the final head crop bounding box. Furthermore, we downscale the head bounding box by 10\% and resize it to $224 \times 224$ pixels. We show the impact of the head bounding box size in the supplementary materials. An 8-frame head crop clip is used for the video modalities, and the frame rate is unified across all video datasets.
All the backbones used in this work are pre-trained on Imagnet for static backbones and ImageNet-1K, Kinetics-400, and SUN RGB-D for Swin3D. Please refer to the supplementary materials for training and data augmentation details.

\vspace{-2mm}
\subsection{Ablation Study}
\vspace{-2mm}
\mypartitle{Does gaze following label improve 3D gaze estimation?}
In \cref{tab:ablation_framework}, we evaluate various methods for learning from 2D gaze following labels (GF). We find that, with few exceptions, incorporating GF consistently improves 3D gaze estimation. This underscores the importance of broader training diversity for robust 3D gaze estimation. The specific details and advantages of each approach are discussed in the following section.\\
\mypartitle{Self-Training Weakly-Supervised learning framework.} In \cref{tab:ablation_framework}, we perform ablation experiments related to the learning framework. In our experiments, we train with our model \model on two 3D gaze datasets namely G360 and GFIE. The first baseline experiment is to train on a 3D gaze dataset in a standard supervised manner. Then, there are three possibilities to leverage additional 2D datasets such as GF. The first one is the Weakly-Supervised (WS) method defined by a specific loss applied only on the x, y 3D gaze prediction coordinate for GF batch samples supervised by the 2D GT label. The second approach Self-Training (ST) is similar to our \framework approach described in \cref{fig:approach} but in the second stage, 3DPred is used to supervise the training. Finally, the last approach \framework is our proposed approach described in \cref{fig:approach} and \cref{sec:self_taining}. 
Compared to ST, \framework achieves higher accuracy across all evaluations except when trained on G360 and tested on EDIAP. This indicates that relying solely on 3DPred lacks diversity in gaze distribution, as it mostly follows the training data distribution. By incorporating our 3DGP label, we obtain a more robust gaze vector that enhances accuracy. 
Furthermore, when compared to using only 2D labels in the WS method, \framework performs slightly better overall, particularly on the frontal EDIAP and MPII benchmarks. This suggests that datasets, where the z component is significant (e.g., frontal views), require more than 2D supervision. Overall, our method is either the best or the second best (by a small difference) demonstrating the effectiveness of our \framework learning framework. \\
\mypartitle{Gaze Model Network.}
In \cref{tab:ablation_model}, we present the results concerning the model architecture. In our experiments, we train on G360 in a supervised manner with different training modalities combinations ( I, V, and I\&V). We compare our model \model with baseline models (see \cref{sec:model}) that can also handle different training data modalities. First, we notice that when models are trained on both image and video, our \model model is the best model on G360. It suggests that spatiotemporal learning from the input is beneficial for gaze estimation, especially in non-frontal scenarios. Additionally, within each model, training on both modalities improves image evaluation but slightly reduces video evaluation. However, modality-specific models are limited to their own data type which limits the available training dataset. Indeed, in the next section, we show that combining modalities can result in cross-modal generalization. Given these findings, our model stands out as a reliable and versatile option for robust 3D gaze estimation.
\\
\mypartitle{Does training on additional image datasets help video generalization?} 
As observed in \cref{tab:ablation_framework}, training our \model model with \framework, which includes a diverse image dataset (GF), not only improves image generalization but, more notably, enhances generalization on videos. Compared to a supervised method, our approach improves image GFIE evaluation by 38\% and video GFIE by 34\%. A similar trend is observed when training on GFIE and evaluating on G360 Full. 
Furthermore, the modality of the training data plays an important role, as observed in \cref{tab:ablation_cross_modal}. Indeed, when training with both an image and a video dataset such as G360V+GF, at test time, evaluation on images (respectively videos) will be dominantly affected by the characteristics of the image (respectively video) training data. When we train on G3360V+GF, the results on GFIE image have a 15.7° angular error against 17.9° on video. Interestingly, when trained on G360I\&V+GF, the generalization modality gap is reduced with 15.9° on image and 15.5° on video. A Similar trend is observed when training on GFIE.  

\mypartitle{Temporal Context.} 
Temporal dynamic plays a crucial role in unconstrained gaze estimation, as evidenced in \cref{tab:ablation_framework} with our \model model trained on both modalities (I\&V).
In all configurations, video predictions consistently outperform image-based predictions.
Other important observations emerged from visual and quantitative analyses and are discussed in the supplementary material.

\begin{table*}[t]
\centering
\resizebox{\textwidth}{!}{%
\begin{tabular}{llrrrrrrrrrrrrrrr|rr}
\toprule
 &  & \multicolumn{2}{c}{G360 Full} & \multicolumn{1}{l}{} & \multicolumn{2}{c}{G360 180} & \multicolumn{1}{l}{} & \multicolumn{2}{c}{G360 40} & \multicolumn{1}{l}{} & \multicolumn{2}{c}{G360 Face} & \multicolumn{1}{l}{} & \multicolumn{2}{c}{G360 Back} & \multicolumn{1}{}{} & \multicolumn{2}{c}{GFIE} \\ \cline{3-4} \cline{6-7} \cline{9-10} \cline{12-13} \cline{15-16} \cline{18-19} 
Method  & Training Data & Img & Vid &  & Img & Vid &  & Img & Vid &  & Img & Vid &  & Img & Vid &  & Img & Vid \\
\midrule
Gaze360~\cite{Gaze360}  & G360I & 15.6 & - &  & 13.4 & - &  & 13.2 & - &  & - & - &  & - & - &   & - & - \\
Kothari \etal~\cite{kothari2021weakly} & G360I & 15.07 & - &  & - & - &  & 10.94 & - &  & - & - &  & - & - &   & - & - \\
MSA~\cite{chen2020360}  & G360I & 13.9 & - &  & 12.2 & - &  & - & - &  & - & - &  & 23.5 & - &   & - & - \\
Gaze360~\cite{Gaze360} & G360V & - & 13.5 &  & - & 11.4 &  & - & 11.1 &  & - & - &  & - & - &   & - & - \\
Kothari \etal~\cite{kothari2021weakly}  & G360V & - & 13.2 &   & - & - &   & - & 10.1 &   & - & - &   & - & - &   & - & -  \\
Kothari \etal~\cite{kothari2021weakly}  & G360V+AVA & - & 13.2 &   & - & - &   & - & 10.2 &   & - & - &   & - & - &   & - & -  \\
MCGaze~\cite{guan2023end}  & G360V & - & 12.96 &  & - & 10.74 &  & - & 10.02 &  & - & - &  & - & - &   & - & - \\
MSA+Seq~\cite{chen2020360}  & G360V & - & 12.5 &  & - & 10.7 &  & - & - &  & - & - &  & - & 19.0 &   & - & - \\
\hline
 Supervised (\model) & G360I\&V & 13.64 & 12.60 &   & 11.66 & 10.67 &   & 9.10 & 8.23 &   & 11.20 & 10.25 &   & 20.74 & 19.53 &   & 21.86 & 20.89 \\
 \framework (\model) & G360I\&V+GF & \textbf{13.19} &\textbf{ 12.17} &   & \textbf{11.34} & \textbf{10.35} &   &\textbf{ 8.58} & \textbf{7.67} &   & \textbf{10.84} & \textbf{9.92} &   & \textbf{19.82} & \textbf{18.72} &   & \textbf{15.90} & \textbf{15.51} \\
\midrule 
\midrule 
GFIE~\cite{GFIE} & GFIEI & - & - &   & - & - &   & - & - &   & - & - &   & - & - &   & 17.7 & - \\
Kothari \etal\cite{kothari2021weakly}   & ETH & 52.6 & - &   & - & - &   & 20.5& - &   & - & - &   & - & - &   & - & - \\
Kothari \etal\cite{kothari2021weakly}  & ETH+AVA& - & 25.0 &   & - & - &   & - & 16.9 &   & - & - &   & - & - &   & - & - \\
3DGazeNet~\cite{ververas20223dgazenet}$^\dag$ & ETH & - & - &   & - & - &   & - & - &   & \textcolor{gray}{22.1}$^\dag$ & - &   & - & - &   & - & - \\
3DGazeNet~\cite{ververas20223dgazenet}$^\dag$  & ETH+AVA+CMU& - & - &   & - & - &   & - & - &   & \textcolor{gray}{17}$^\dag$ & - &   & - & - &   & - & - \\
3DGazeNet~\cite{ververas20223dgazenet}$^\dag$  & ETH+ITWG-MV& - & - &   & - & - &   & - & - &   & \textcolor{gray}{15.4}$^\dag$ & - &   & - & - &   & - & - \\
\hline
 Supervised (\model) & GFIEI\&V & 30.57 & 29.90 &   & 29.08 & 28.65 &   & 33.43 & 32.94 &   & 28.87 & 28.56 &   & 35.95 & 34.40 &   & 15.70 & 15.44 \\
 \framework (\model) & GFIEI\&V+GF & \textbf{21.48} & 21.06 &   & 20.61 & 20.32 &   & 26.55 & 26.73 &   & 20.46 & 20.23 &   & \textbf{24.61} & \textbf{23.75} &   & 12.99 & 12.68 \\
 Supervised (\model) & GFIEI\&V+MPS & 25.75 & 24.29 &   & 20.35 & 19.42 &   & 16.35 & 15.49 &   & 19.07 & 18.28 &   & 45.19 & 41.81 &   & 15.61 & 15.38 \\
 \framework (\model) & GFIEI\&V+MPS+GF & 21.59 & \textbf{20.02} &   & \textbf{17.02} & \textbf{15.67} &   & \textbf{13.90} & \textbf{12.66} &   & \textbf{16.00} & \textbf{14.75} &   & 38.02 & 35.69 &   & \textbf{12.82} & \textbf{12.49} \\
\bottomrule
\end{tabular}%
}
\vspace{-2mm}
\caption{ \textbf{Comparison with the state-of-the-art on physically unconstrained benchmark Gaze360 and GFIE test set.} We report both within and cross-dataset evaluation trained using \model model with and without our \framework framework.
The top table presents methods trained with Gaze360, while at the bottom, methods are trained using GFIE. The method with $\dag$ is restricted with only frontal pose with face and eye crop as input. Moreover, the method is evaluated on a new subset (head pose yaw $\in$ [-90,90]), which is close to, but not the same as, the Gaze360 Face subset.
}
\label{tab:main_results}
\vspace{-4mm}
\end{table*}
\begin{table}[t]
\centering
\resizebox{\columnwidth}{!}{%
\begin{tabular}{llrrrrrrrr}
\toprule
 &  & \multicolumn{2}{c}{G360 Face} & \multicolumn{1}{l}{} & \multicolumn{2}{c}{G360 Face 180} & \multicolumn{1}{l}{} & \multicolumn{2}{c}{G360 Face 40} \\ \cline{3-4} \cline{6-7} \cline{9-10} 
Method & Training Data & Img & Vid &  & Img & Vid &  & Img & Vid \\ 
\midrule
FullFace~\cite{zhang2017s}              & G360I Face & 14.99 & - &  & - & - &  & - & - \\
Dilated~\cite{chen2018appearance}       & G360I Face & 13.73 & - &  & - & - &  & - & - \\
RT-Gene~\cite{Rt-gene}                  & G360I Face & 12.26 & - &  & - & - &  & - & - \\
CA-Net~\cite{cheng2020coarse}           & G360I Face & 11.20 & - &  & - & - &  & - & - \\
Gaze360~\cite{Cheng2021Survey}          & G360V Face & - & 11.04 &  & - & - &  & - & - \\
ResNet50~\cite{cheng2022gaze}           & G360I Face & 10.73 & - &  & - & - &  & - & - \\
GazeTR~\cite{cheng2022gaze}             & G360I Face & \textbf{10.62} & - &  & - & - &  & - & - \\
L2CS~\cite{abdelrahman2023l2cs}         & G360I Face & - & - &  & 10.4 & - &  & 9.0 & - \\
SPMCCA~\cite{yan2023gaze}               & G360I Face & - & - &  & \textbf{10.16} & - &  & 8.62 & - \\
SAM-LSTM~\cite{jindal2024spatio}         & G360V Face & - & 10.05 &  & - & 9.84 &  & - &  \textbf{6.92} \\
\midrule
Supervised (\model) & G360I\&V    & 11.20 & 10.25 &   & 11.01 & 10.09 &   & 8.81 & 8.02 \\
\framework (\model) & G360I\&V+GF    & 10.84 & \textbf{9.92} &   & 10.65 & \textbf{9.75} &   & \textbf{8.30} & 7.48 \\
\bottomrule
\end{tabular}%
}
\vspace{-2mm}
\caption{\textbf{Comparison with the state-of-the-art constrained methods tested on Gaze360 detected face}. All the state-of-the-art methods use a face crop as input and are trained on the detected face subset of Gaze360. We report results trained on G360 Full using \model and with or without \framework. Note that the other methods are constrained to face input therefore our method is more general and can be applied to any head pose orientation. }
\vspace{-5mm}
\label{tab:within_constraint}
\end{table}

\vspace{-1mm}
\subsection{Comparison with State-of-the-art (SOTA).}
\vspace{-1mm}
\mypartitle{Within-Dataset Experiments.}
In the following, we focus on within-dataset experiments.
In \cref{tab:main_results}, we compare our results with the state-of-the-art methods on G360 and GFIE. We report results using our model \model in a supervised manner and using our \framework learning framework. Our model trained in a supervised manner is SOTA on image G360 Full and GFIE with 2\% and 13\% relative improvement, respectively. On video inference, MSA+Seq is slightly better (12.6° vs 12.5° ours) since it uses an average of multiple input scales.
More importantly, when trained with gaze following labels like GF using our \framework learning framework, we outperform all the SOTA on image and video by 5\% on G360 Full image, 3\% on G360 Full video, and 36\% on GFIE image. In contrast, Kothari \etal~\cite{kothari2021weakly} don't improve when using the LAEO label (AVA) in a weakly-supervised framework. 
Additionally, in \cref{tab:within_constraint}, we compared our method trained on G360 Full to methods trained on detected face subset G360 Face. Given that the state-of-the-art methods are specifically designed for near-frontal faces, our supervised model \model is not SOTA but demonstrates competitive performance. When including gaze following label using our \framework framework, it shows very competitive results and SOTA performance on G360 Face video (9.92° vs 10.05°), G360 Face 180 video (9.84° vs 9.75°), and G360 Face 40 (8.62° vs 8.30°). Therefore, compared to methods using tight face crops (increasing eye resolution), our \framework approach proved to be competitive on near frontal view. We include a comparison with SOTA trained on G360 and evaluated on constrained benchmarks MPII and EDIAP in supplementary materials. \\
\mypartitle{Cross-Dataset Experiments. }
In this section, we emphasize on cross dataset experiments. In \cref{tab:main_results}, we compare our method with SOTA methods on generalization on G360 (bottom part).
Among the few approaches that explore generalization on Gaze360, Kothari \etal~\cite{kothari2021weakly} provides the most relevant comparison. In contrast, 3DGazeNet~\cite{ververas20223dgazenet} provides cross-dataset generalization on G360 Face but only works on frontal faces requiring face and eye crop as input.  
Our results show that our \framework framework trained with various available 3D gaze datasets (GFIE, or GFIE+MPS) always improves generalization. For instance, when tested on G360 Full image and video, it always outperforms our supervised approach by 40\% and 20\% when trained with GFIE or GFIE+MPS, respectively. A similar trend is observed when trained on G360 and tested on GFIE. Therefore, it confirms that our framework using gaze follow labels is effective for improved generalization. \\
When compared to Kothari \etal~\cite{kothari2021weakly} trained on LAEO labels (AVA), our \framework approach trained using GFIE+GF shows better performance on G360 Full but is behind on G360 40 because GFIE doesn't contain frontal samples. In contrast, when trained using GFIE+MPS+GF, it improves over Kothari on both G360 Full and 40.\\
\mypartitle{Limitations. }
As expected when using our framework, the generalization improvement is tight to the training diversity used in the pre-training stage. In cross-dataset experiments in \cref{tab:main_results}, compared to our supervised model, we observe that when trained using GFIE our framework improves more on G360 Back and less on G360 40 because of the non-frontal distribution of GFIE.

\vspace{-2mm}
\section{Conclusion}
\vspace{-1mm}

In this work, we introduced the \framework learning framework, which leverages weakly annotated images with 2D gaze datasets, such as gaze-follow labels, to enhance appearance diversity and broaden gaze distributions across natural scenes. We also presented our Gaze Transformer, \model, which improves performance and supports both image and video training. By combining \framework and \model, we achieve significant gains in both within- and cross-dataset experiments, reaching state-of-the-art results on GFIE and Gaze360. Additionally, we demonstrate effective cross-modal generalization, a critical capability given the scarcity of video datasets. We believe our approach is a promising solution for robust 3D gaze tracking in the wild, suitable for a range of challenging applications.

\clearpage
{
    \small
    \bibliographystyle{ieeenat_fullname}
    \bibliography{reference}
}



\twocolumn[{%
 \centering
 \LARGE \papertitle \\[1em]
 \large Supplementary Material\\[1em]
}]


\renewcommand*{\thesection}{\Alph{section}}
\setcounter{section}{0}
\setcounter{equation}{0}
\setcounter{figure}{0}
\setcounter{table}{0}
\makeatletter
\renewcommand{\theequation}{S\arabic{equation}}
\renewcommand{\thefigure}{S\arabic{figure}}
\renewcommand{\thetable}{S\arabic{table}}


\mypartitle{What is expected?} The supplementary material consists of datasets details, experiments details, and extended experiments analysis mentioned in the main paper. In addition, videos of qualitative examples of our method on VideoAttentionTarget further demonstrate the robustness in challenging real-world scenarios.  

\section{Datasets Details}

\subsection{Datasets}
\label{sec:supp_dataset}
\mypartitle{ Gaze360 (G360).} \cite{Gaze360} is video 3D gaze datasets. It is collected in both indoor and outdoor environments in unconstrained setting, which contains 3D gaze of 238 subjects with a wide-range head pose and gaze direction. G360 is recorded at 8FPS. In all of our experiments, we always used the same training set as \cite{Gaze360} with 126928 samples. For the test set, we followed the split of \cite{Gaze360} where G360 Full corresponds to "All 360°" (the entire test set) with 25969 samples, G360 180 corresponds to "Front 180°" (gaze within 90°) with 20322 samples, and G360 40 to "Front Facing" (gaze within 20°) with 3995 samples. In addition to those splits, we consider G360 Back (gaze above 90°) \cite{chen2020360} with 5647 samples and finally G360 Face (all detected faces) with 16031 samples, which is used in many constrained gaze studies \cite{zhang2017s,chen2018appearance,Rt-gene,cheng2020coarse,cheng2022gaze,abdelrahman2023l2cs,yan2023gaze,catruna2024crossgaze}. When we refer to G360 Face 180 (15895 samples), it corresponds to the detected face with a gaze within 90°, a subset of G360 180, the same for G360 Face 40 with 3687 samples. We used the validation set described in \cite{Gaze360} with 17038 samples.

\mypartitle{GFIE.} \cite{GFIE} is a video 3D gaze dataset collected indoors with 71799 frames from 61 subjects (27 male and 34 female). It is an unconstrained dataset with a wide range of head poses. It was collected for gaze following task; using a complex calibrated laser setup, they can infer the 3D gaze from the eye to the visual target direction. They recorded people doing various indoor activities at 30 fps. We follow the data splits described in \cite{GFIE}, 59217 for training, 6281 for validation, and 6281 for testing.

\mypartitle{MPSGaze (MPS).} \cite{MPS} is a modified 3D gaze datasets that has been automatically generated using ETH-Xgaze \cite{ETH} eyes. They apply a blending technique on people from the Widerface~\cite{yang2016wider} dataset to put eyes with a known 3D gaze from ETH on heads with similar head poses. This dataset is diverse, with more than 10k identities and challenging poses, appearances, and lighting conditions. However, the blending process reduces the quality of the visual appearance, and it contains only near frontal head poses and no back view. We used the same training and test split with 24282 samples in training and 6277 samples in testing. No validation is defined in this work. 

\mypartitle{EYEDIAP (EDIAP).} \cite{Eyediap} is a 3D gaze video dataset. It includes videos from 16 subjects (30 fps), using either screen targets (CS, DS subset EDIAP) or 3D floating balls ( FT subset EDIAP-FT) as gaze targets. It is a constrained setup with mainly frontal head poses. 
Following \cite{wang2022contrastive,Cheng2021Survey}, we used the evaluation set under screen target session (CS, DS, namely EDIAP) with 16674 samples from 14 subjects. 

\mypartitle{MPIIFaceGaze (MPII).} \cite{MPIIGaze} is a 3D gaze image dataset collected from 15 subjects in a screen-based gaze target setup, resulting in a constrained dataset with mostly frontal head pose. We follow the standard evaluation protocol \cite{MPIIGaze,wang2022contrastive,Cheng2021Survey}, which selects 3000 images from each subject to form an evaluation set for a total of 45000 samples. 

\mypartitle{GazeFollow (GF).} \cite{recasens2015they} is a 2D gaze image dataset annotated on in the wild dataset for the gaze the following task. The 2D target label corresponds to where a given person is looking at in the image. It is a diverse dataset that includes various head poses, appearances, scenes, and lighting conditions. Overall, it has around 130K annotated person-target instances in 122K images.

\subsection{Video Processing}
As mentioned in the main section, for video clip input, our approach predicts the 3D gaze from an 8-frame video clip. However, video datasets have different frame rates, which can impact the gaze prediction. In this work, since G360 has a lower frame rate, we resample EYEDIAP and GFIE to match G360's frame rate of 8 fps. 

\subsection{Gaze Representation} 
Working with different 3D gaze datasets requires a unified way to define and represent the 3D gaze vector. Usually, in constrained gaze estimation, studies use data normalization to map the input image to a normalized space where a virtual camera is used to warp the face patch out of the original input image according to the 3D head pose \cite{ETH}. Thus, the gaze is expressed in this virtual camera coordinate defined by the 3D head pose. \\
However, in unconstrained settings, it is not possible to get access to a robust and reliable 3D head pose; thus, we follow the gaze representation of Gaze360 \cite{Gaze360} in the ``Eye coordinate system". The practical interpretation of the eye coordinate system is that the positive x-axis points to the left, the positive y-axis points up, and the positive z-axis points away from the camera, \ie [-1,0,0] is a gaze looking to the right or [0,0,-1] straight into the camera from the camera's point of view, irrespective of subjects position in the world. The origin of the gaze vector is the middle of the eyes, except for MPS and MPII, where the gaze origin is the average of 3D eyes and mouth landmarks resulting in an origin located at the middle of the nose, and for GF, we used the center of the head bounding box as the origin.

\begin{figure}
    \centering
    \includegraphics[width=0.99\linewidth]{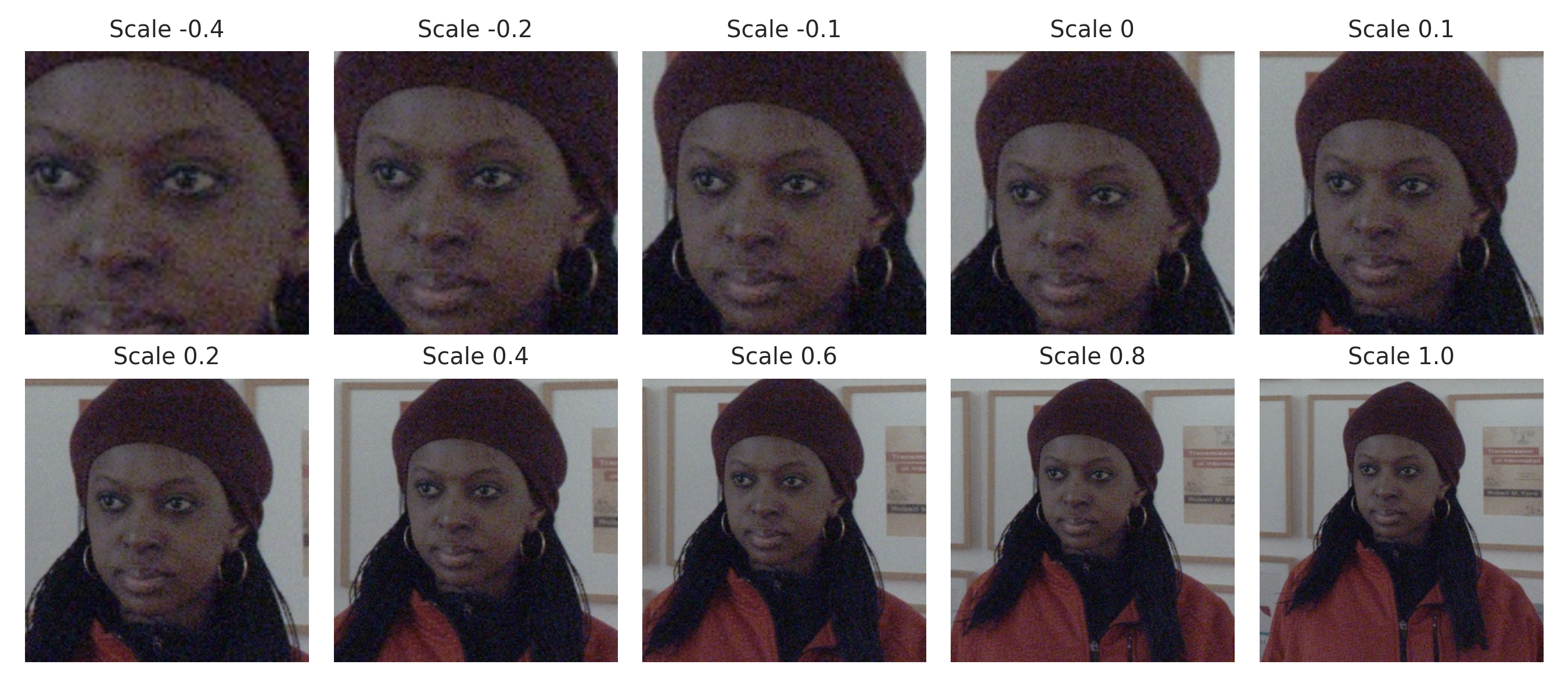}
    \vspace{-4mm}
    \caption{Input head crop using different scales. In our work, a scale of -0.1 is used and proved to be effective in both constrained and frontal face setting \cref{sec:crop_size} }
    \label{fig:head_crop_input}
\end{figure}

\section{Experiments Details}

\mypartitle{Metric.}
We follow the test split described in the state-of-the-art method and explained in \cref{sec:supp_dataset}. As a metric, we use the standard angular error in degrees between the predicted and ground truth gaze prediction \cite{Gaze360,ETH,Eyediap,MPIIGaze}. Previous methods reporting video evaluation used a 7-frame video clip and predict the middle frame gaze direction. Since our approach outputs eight gaze directions from an 8-frame video clip, for a fair comparison, we use the 4th gaze prediction of an 8-frame video clip to compute the evaluation metric. 

\mypartitle{Training.}
We used the same setup in all the experiments to be as fair as possible. All the models are trained for a minimum of 20 epochs. We used an early stopping on the validation set with a patience of 10 epochs. We use the AdamW optimizer \cite{AdamW} with a learning rate of 1e-4 and a cosine annealing schedule with a 5 epochs linear warmup (from 2e-5 to 1e-4). For evaluation, we report the performance of the best model defined by the best angular error on the validation set.

\mypartitle{Data augmentation.}
Data augmentation is crucial for robust gaze estimation in the wild. In this work, we used standard data augmentation techniques. First, we applied jittering during the head crop to introduce slight variations in scale and aspect ratio, which reduces the model's sensitivity to noisy or imprecise head bounding boxes. Next, color jittering was applied by adjusting brightness, contrast, and saturation, making the model more resilient to diverse lighting conditions commonly encountered in real-world scenarios. Since gaze labels, such as those in the GF 2D dataset, may exhibit bias toward one side, we applied horizontal flipping to the images while appropriately adjusting the gaze direction, ensuring more balanced training data in the yaw gaze direction. These augmentations collectively improved the model’s ability to handle variations in data and enhance its generalization to unseen environments.

\section{Additional Experiments}

\subsection{Effect of Head Crop Size}
\label{sec:crop_size}
\begin{figure}[htbp]
    \centering
    \begin{subfigure}{\linewidth}
        \centering
        \includegraphics[width=0.99\textwidth]{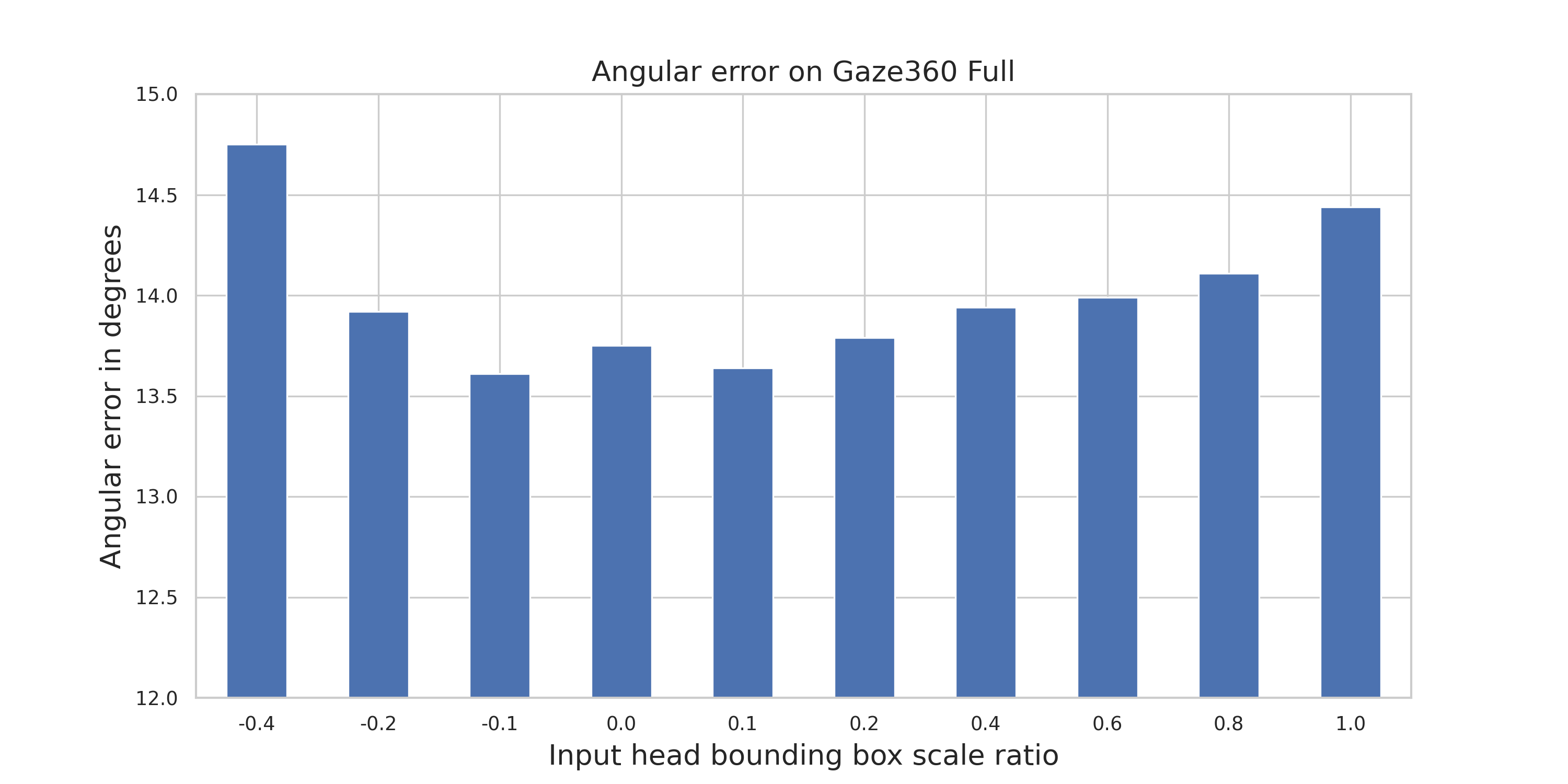}
        \caption{Effect of head bounding box scale as input on the 3D gaze angular error on G360 Full test set. A scale ratio of 0.1 corresponds to a 10\% bounding box scale.}
        \label{fig:supp_head_crop_angular}
    \end{subfigure}

    \begin{subfigure}{\linewidth}
        \centering
        \includegraphics[width=0.8\textwidth]{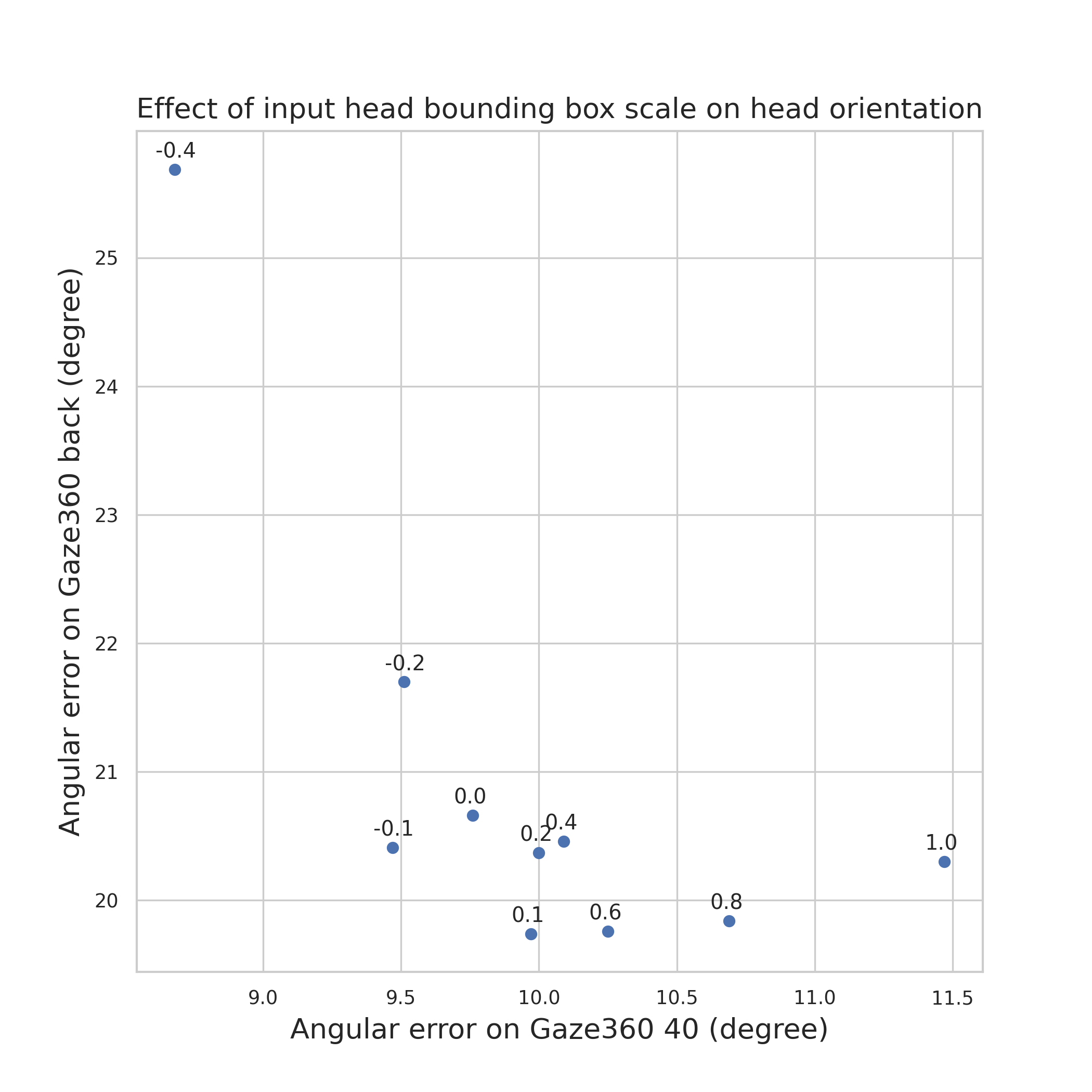}
        \caption{Effect of head bounding box scale on the angular error with respect to G360 Back and G360 40 test subset.}
        \label{fig:supp_head_crop_effect}
    \end{subfigure}
    
    \caption{\textbf{Effect of head crop size.}}
    \label{fig:head_crop}
\end{figure}

As mentioned by Chen \etal~\cite{chen2020360}, the input head crop scale impacts the 3D gaze estimation. We find that the effect on the prediction depends on the head orientation. \cref{fig:head_crop_input} illustrates the different inputs with different head crop scales. As shown in \cref{fig:supp_head_crop_effect}, a smaller head crop tighter to the face improves 3D gaze estimation on frontal head poses, while a larger head crop improves gaze on the non-frontal head pose. Indeed, as shown in \cref{fig:head_crop_input}, a tighter crop increases the eye resolution in the image and a larger crop provides more context about the head orientation and upper body orientation, which gives a strong prior for the gaze direction when eyes are not visible.
In the context of gaze estimation in the wild, a scale of -10\% is part of the Pareto front as illustrated in \cref{fig:supp_head_crop_effect} and is also the best on the G360 Full image as shown in \cref{fig:supp_head_crop_angular}. Therefore, it is a reasonable trade-off between frontal and back view performance. We use it for all our experiments.

\subsection{Constrain Gaze Evaluation}

\begin{table}[t]
\centering
\resizebox{\columnwidth}{!}{%
\begin{tabular}{llrrrrrrr}
\toprule
 &  & \multicolumn{1}{c}{MPII} & \multicolumn{1}{l}{} & \multicolumn{2}{c}{EDIAP}  \\ \cline{3-3} \cline{5-6} 
Method & Training Dataset & Img &  & Img & Vid  \\ 
\hline
PureGaze~\cite{cheng2022puregaze} (Res18)        & G360I Face & 9.3 &   & 9.2 & -  \\
Liu \etal~\cite{liu2021generalizing} (Res18)      & G360I Face & 7.7 &   & 9.0 & -  \\
Liu \etal~\cite{liu2021generalizing} (Res50)     & G360I Face & 8.3 &   & 7.5 & - \\
RAT~\cite{bao2022generalizing} (Res18)     & G360I Face & 7.6 &   & \textcolor{gray}{7.1}$^*$ & -  \\
RAT~\cite{bao2022generalizing} (Res50)     & G360I Face & 7.7 &   & \textcolor{gray}{7.1}$^*$ & -  \\
CDG~\cite{wang2022contrastive} (Res50)           & G360I Face & 7.0 &   & \textbf{7.3} & -  \\
\midrule
 Supervised (\model) & G360I\&V             & 7.43 &   & 8.88 & 8.28  \\
 \framework (\model) & G360I\&V+GF          & \textbf{6.43} &   & 8.87 & 8.19  \\
\bottomrule
\end{tabular}%
}
\caption{\textbf{Comparision with state-of-the-art on constrained domain generalization benchmarks}. All these methods \cite{cheng2022puregaze,liu2021generalizing,wang2022contrastive,ververas20223dgazenet,bao2022generalizing} use a face crop as input and are trained on the detected face subset of Gaze360. Our method is trained and tested on head crop which makes it more general but more challenging for frontal gaze estimation. $^*$ In \cite{bao2022generalizing} they used only 6400 sample for EDIAP but we follow \cite{Cheng2021Survey,cheng2022puregaze,wang2022contrastive} with 16674 samples. }
\label{tab:generalization_constrained}
\vspace{-3mm}
\end{table}
\vspace{-1mm}

The objective of this work is to improve unconstrained gaze estimation in the wild. As seen in \cref{sec:crop_size}, compared to a tight face crop a larger crop improves gaze in challenging head pose. Therefore, a larger crop is more suited to our objective.  In contrast, some methods specialize in frontal gaze estimation and rely on tight face crops, which provide better resolution for the eye regions. While this is not a fully fair comparison, we compare our approach to these constrained methods for generalization on constraint benchmarks. Note that for the constrained methods, models are trained and tested only on a subset of detected faces (G360 Face), while in our approach the model is trained on G360 Full. \\
As shown in \cref{tab:generalization_constrained}, on MPII, the supervised GaT lags behind the best method by 6\%. On EDIAP, GaT is 21\% behind the best method in image evaluation but narrows the gap to 13\% when evaluated on videos. Then, when using our \framework learning framework including GF labels, we observe an important improvement on MPII with state-of-the-art angular error of 6.43 compared to 7 from CDG. On EDIAP the improvement is marginal. Compared to EDIAP, MPII has more diversity in lighting conditions and environment. GF doesn't contain a lot of frontal gaze direction but has a broad diversity of environments. Therefore, the improvement on MPII should come from the additional diversity that GF brings but this is not useful for EDIAP prediction. 
While constrained methods excel in frontal settings, they fail in unconstrained scenarios. Our approach, which achieves state-of-the-art performance in unconstrained environments (G360, GFIE) while remaining competitive in constrained settings (MPII, EDIAP), proves to be a versatile and robust solution for gaze estimation in the wild.

\subsection{Qualitative Analysis}

\begin{figure}[t]
    \centering
    \includegraphics[width=0.8\linewidth]{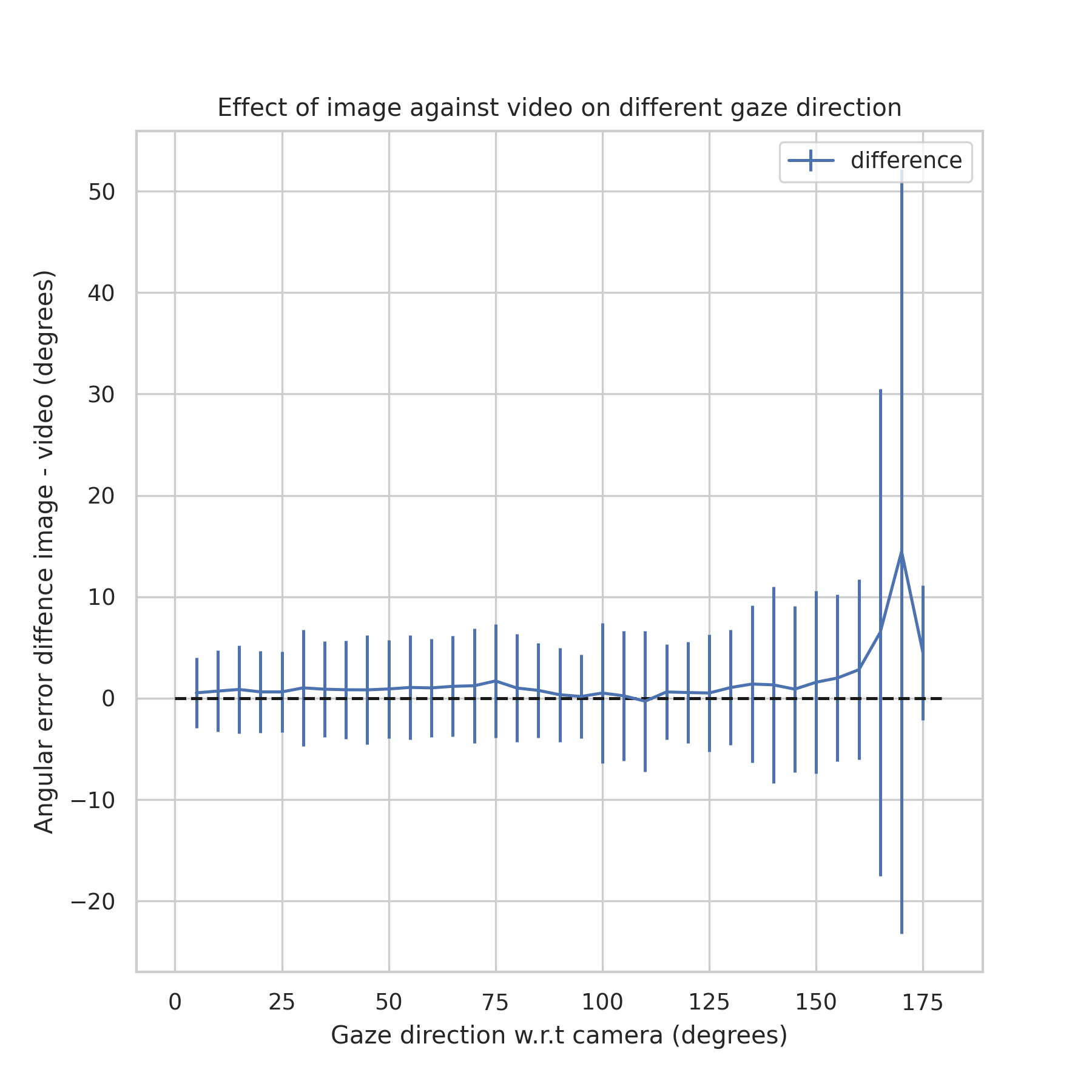}
    \caption{ \textbf{Image vs video predictions, where does it help?}. GaT trained on G360I\&V and tested on G360 Full image and video. The difference between image and video angular error with respect to the ground truth gaze directions from the camera ([0,0,-1]). The mean and standard deviation are displayed for each 10° bin. Positive values indicate better performance in video prediction compared to image prediction. }
    \label{fig:graph_img_vs_video}
\end{figure}

\begin{figure*}[t]
    \centering
    \includegraphics[width=1\linewidth]{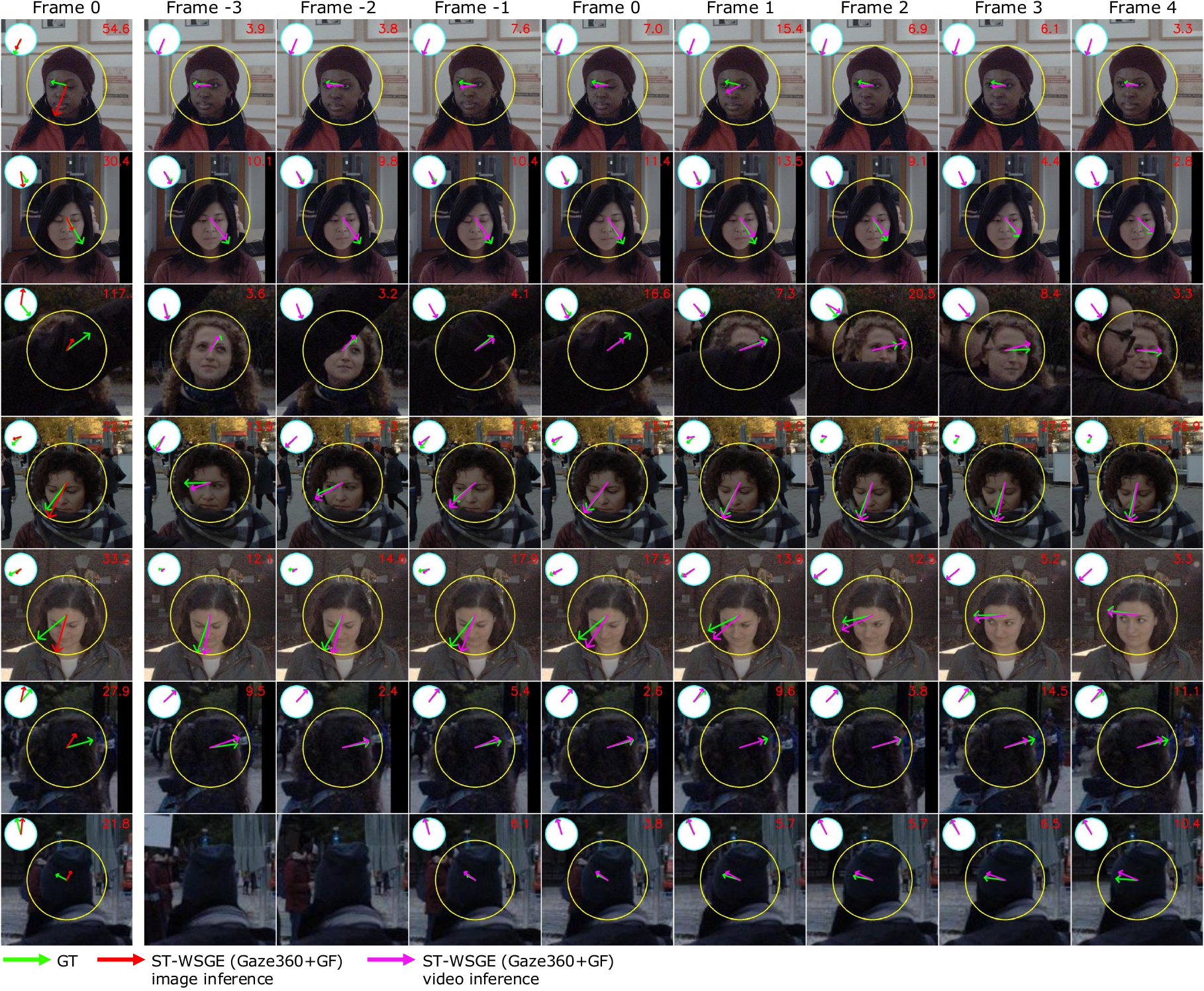}
    \caption{ \textbf{Illustration of image against video prediction.} Comparison between single-image (frame 0) and video predictions (frame -3 to 4). We use our \framework learning framework with \model trained on G360 and GF. All examples are from G360 test set. Rows 1-2 illustrate eye blinks, Row 3 shows an example of occlusion, Rows 4-5 demonstrate frontal head/eyes motion, and Rows 6-7 depict back view prediction. In the last row, the first two frames are not part of the test subset. Arrows in \textcolor{red}{red} represent image predictions, and arrows in \textcolor{magenta}{magenta} are video predictions. The angular error between groundtruth and prediction is displayed in red at the top right corner. The circles in the images represent unit disks where 3D gaze vectors are projected onto the image plane (x,y in yellow) and a top view (x,z in blue)}
    \label{fig:temporal_pros}
\end{figure*}

\begin{figure*}[t]
    \centering
    \includegraphics[width=1\linewidth]{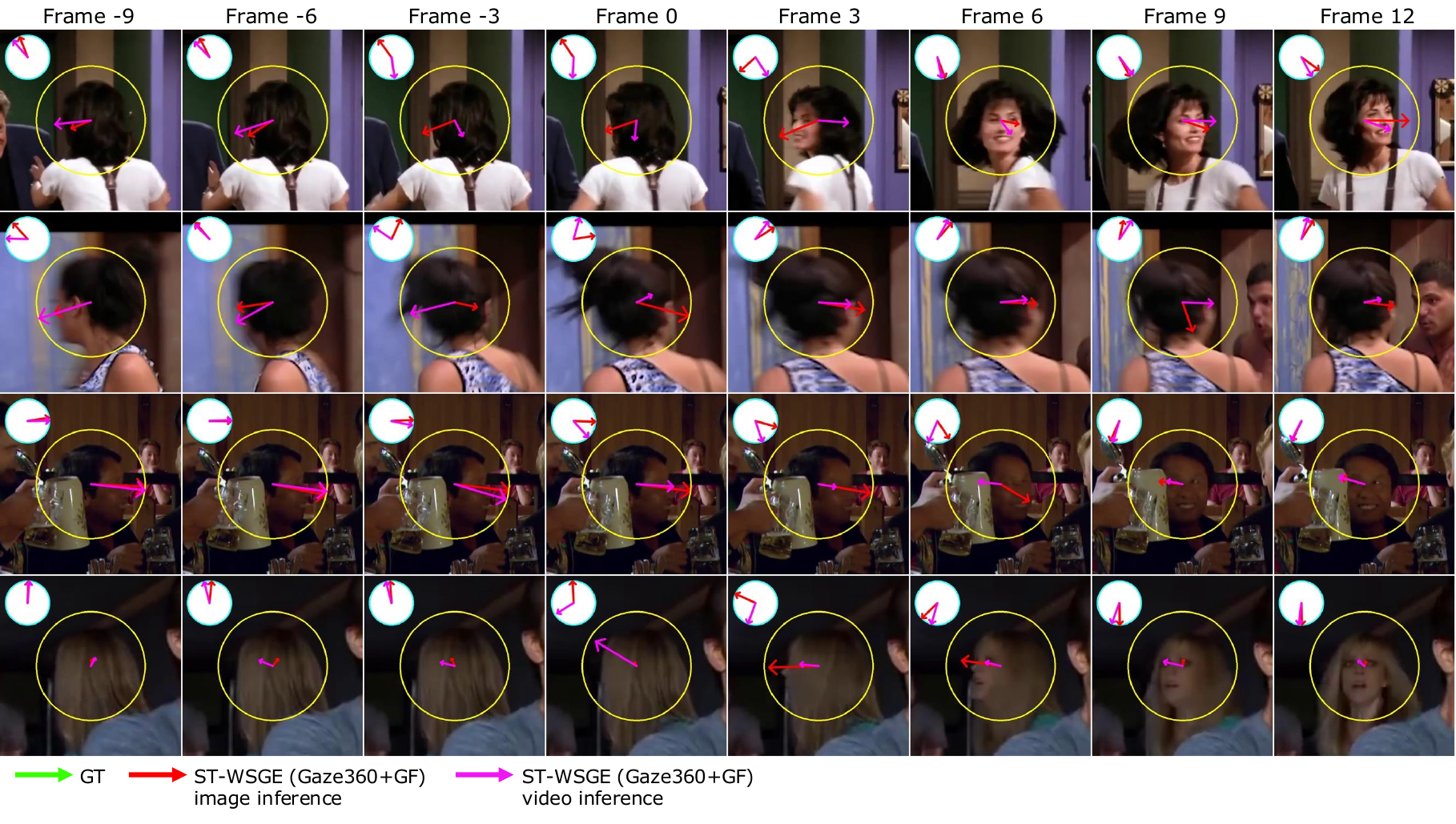}
    \caption{\textbf{Illustration of image and video prediction in case of rapid head motion.} We use our \framework learning framework with \model trained on G360 and GF. All examples are from VideoAttentionTarget~\cite{Chong_2020_CVPR} (VAT). Arrows in \textcolor{red}{red} represent image predictions, and arrows in \textcolor{magenta}{magenta} are video predictions. The circles in the images represent unit disks where 3D gaze vectors are projected onto the image plane (x,y in yellow) and a top view (x,z in blue). Note that since VAT has a frame per second (fps) of 24 and G360 has a fps of 8, we show the temporal context used for video inference corresponding to 8 fps. }
    \label{fig:temporal_cons}
\end{figure*}

\begin{figure*}[t]
    \centering
    \includegraphics[width=1\linewidth]{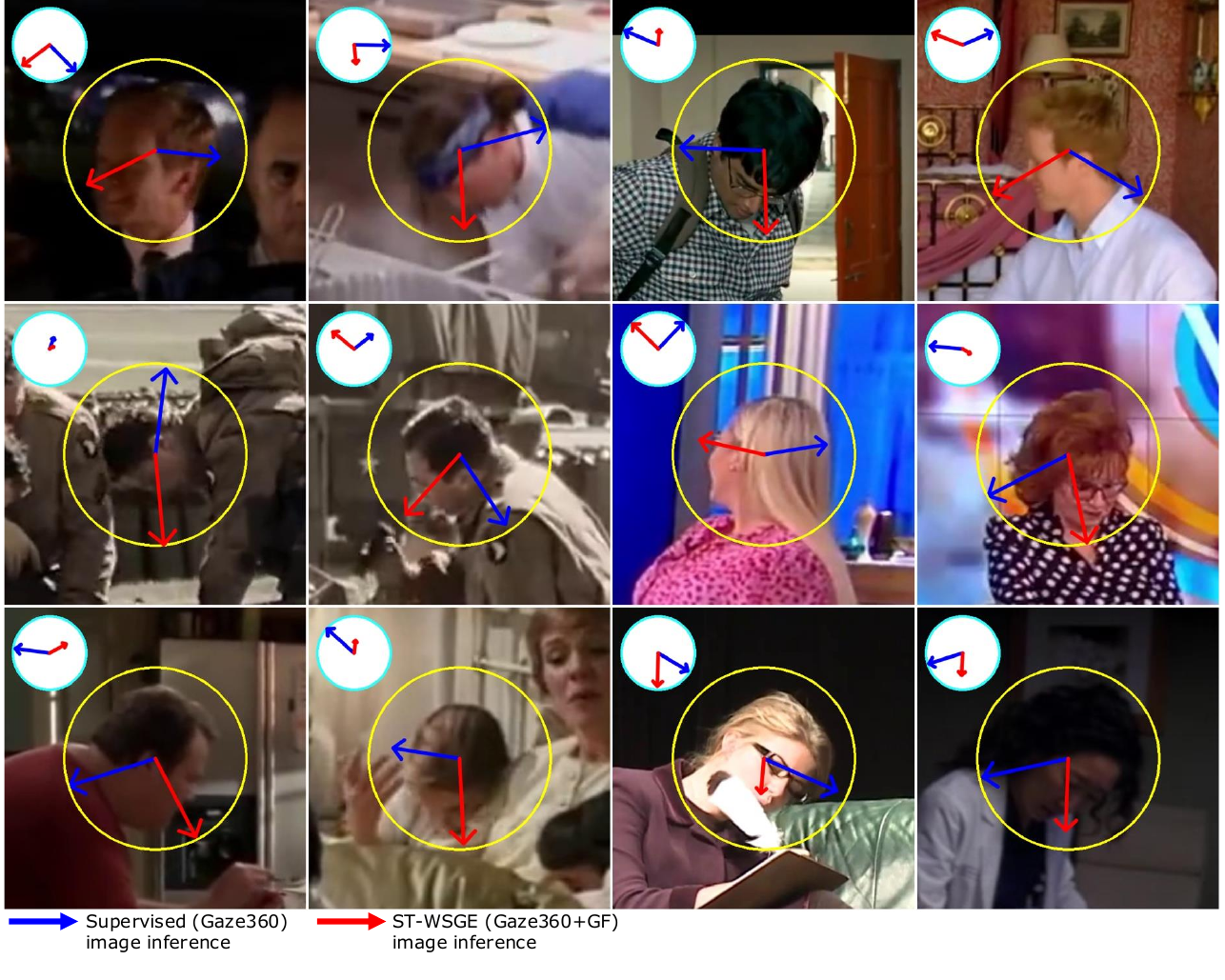}
    \caption{\textbf{Illustration of supervised against \framework learning framework with GazeFollow label.} We use in both experiments our \model model. All examples are from VideoAttentionTarget~\cite{Chong_2020_CVPR} (VAT). Arrows in \textcolor{blue}{blue} represent image predictions with supervised \model trained on G360, and arrows in \textcolor{red}{red} are image predictions with \framework \model trained on G360 and GF. The circles in the images represent unit disks where 3D gaze vectors are projected onto the image plane (x,y in yellow) and a top view (x,z in blue). }
    \label{fig:sup_vs_stwsge}
\end{figure*}

\mypartitle{When does temporal context contribute most effectively?}
As seen in the main paper, video prediction consistently outperforms image prediction. To understand the significance of temporal context in gaze estimation, we examined cases with large angular errors between image and video predictions. Several key observations emerged. As illustrated in \cref{fig:temporal_pros} in the first two rows, temporal context proves valuable during blinks, as it allows the model to interpolate gaze direction when the eyes are closed. If the head pose is not informative, temporal context helps disambiguate between blinking and looking down since the eyes are not visible, as shown in row 1. Additionally, when individuals are viewed entirely from behind (rows 6-7), video inferences provide a more consistent gaze direction in relation to time. Thus, there is less jittering and it might improve the prediction accuracy. In rows 4-5, the head and eye motion can be used in video prediction to improve the gaze direction. Finally, it can help in case of occlusion, as seen in row 3.\\
Furthermore, we explore the impact of image- and video-based prediction with respect to gaze direction. Indeed, we expected more improvement when people are from the back since additional head motion cues can be useful for gaze estimation. In the results, video prediction on G360 Back clearly improves image prediction. In addition, in \cref{fig:graph_img_vs_video}, we plot the difference between image and video prediction angular error for different gaze directions. If we look at the trend, video prediction seems to be better, especially for gaze over 150°, but given the standard deviation, it might not be a statistically significant observation. A more detailed analysis by considering only cases where there is a head motion can better highlight the impact of video prediction.\\
\mypartitle{What are the limitations of temporal context for gaze?}
We investigate prediction made on the VideoAttentionTarget~\cite{Chong_2020_CVPR} (VAT) videos using our \framework framework and \model model. VAT is a challenging dataset with real-world scenarios, various appearances, and diverse gaze distribution, making it well-suited for assessing our approach. Our qualitative analysis reveals two limitations of video-based inference compared to image-based inference using our model. The first limitation arises in cases of rapid head rotation, as illustrated in \cref{fig:temporal_cons}, temporal context may be misused, leading to predictions that do not align with the actual gaze. It might be because no rapid head motion is present in the G360 training sets. 
The second aspect involves cases of ``gaze recentering", where the gaze direction returns to its initial position following a shift. This behavior can occur very rapidly, within just 3-4 frames. Due to the smoothing effect in the temporal modeling, the predicted gaze may not exhibit the same amplitude as the actual movement. Indeed, this behavior is not present in the G360 dataset, and the use of videos sampled at 8 frames per second may limit the ability to capture fine-grained gaze dynamics. However, such behavior is better captured during image-based inference. This highlights a trade-off: while video-based inference provides smoother and more robust predictions, image-based inference offers greater accuracy but can result in jittery outputs. To mitigate the lack of natural gaze behavior we apply our \framework framework using 2D gaze video data from VAT. Unfortunately, since current benchmarks don't contain natural gaze behavior, the results don't show quantitative improvement. Further research to evaluate this aspect is needed.   \\
\mypartitle{In which scenarios does \framework with GazeFollow labels provide the most benefit?}
We demonstrated the advantages of \framework with GazeFollow labels across various benchmarks, both within- and cross-datasets. But in which scenarios does it outperform supervised methods trained solely on G360? To address this question, we analyze predictions made in real-world scenarios using the VideoAttentionTarget (VAT) dataset~\cite{Chong_2020_CVPR}. Our findings reveal that \framework achieves the most notable improvements in cases of extreme head poses, particularly when the head is facing downward, as shown in \cref{fig:sup_vs_stwsge}. It is also more robust to appearance diversity like hair partially occluding the face or varying skin tones. It also helps in difficult lighting conditions and low-resolution inputs.
Additionally, we include a video (provided in the supplementary materials) displaying predictions on VAT with an explanation, enabling a direct comparison between the two methods and a clearer visualization of our approach's performance on real-world data.

\end{document}